\documentclass[11pt]{article}

\PassOptionsToPackage{dvipsnames}{xcolor}
\usepackage[preprint]{acl}

\usepackage{times}
\usepackage{latexsym}
\usepackage[T1]{fontenc}
\usepackage[utf8]{inputenc}
\usepackage{microtype}
\usepackage{inconsolata}
\usepackage{graphicx}

\usepackage{subcaption}
\usepackage{booktabs}
\usepackage{multirow}
\usepackage{amsmath}
\usepackage{amssymb}
\usepackage{mathtools}
\usepackage{amsthm}
\usepackage{xspace}
\usepackage{enumitem}
\usepackage{nicefrac}
\usepackage{xurl}
\usepackage[capitalize,noabbrev]{cleveref}

\newcommand{\model}{\textsc{SWE-Spot}\xspace}

\newcommand{\rayb}[1]{{\textcolor{green}{[Baishakhi: #1]}}}

\newcommand{\magnus}[1]{{\textcolor{cyan}{[Magnus: #1]}}}

\usepackage{pifont}

\title{SWE-Spot: \\Building \underline{S}mall Re\underline{\smash{po}}-Exper\underline{t}s with Repository-Centric Learning}

\author{
 \textbf{Jinjun Peng}\textsuperscript{1}\thanks{Equal contribution.},
 \textbf{Magnus Saebo}\textsuperscript{1}\footnotemark[1],
 \textbf{Tianjun Zhong}\textsuperscript{1},
 \textbf{Junfeng Yang}\textsuperscript{1},
\\
 \textbf{Baishakhi Ray}\textsuperscript{1},
 \textbf{Simin Chen}\textsuperscript{1},
 \textbf{Yangruibo Ding}\textsuperscript{2}
\\
 \textsuperscript{1}Columbia University \quad
 \textsuperscript{2}University of California, Los Angeles
\\
 \small{
   \textbf{Correspondence:} \href{mailto:jinjun@cs.columbia.edu}{jinjun@cs.columbia.edu}, \href{mailto:m.saebo@columbia.edu}{m.saebo@columbia.edu}, \href{mailto:yrbding@cs.ucla.edu}{yrbding@cs.ucla.edu}
 }
}

\begin{document}
\maketitle




\begin{abstract}

The deployment of coding agents in privacy-sensitive and resource-constrained environments drives the demand for capable open-weight Small Language Models (SLMs).
However, they suffer from a fundamental capability gap: unlike frontier large models, they lack the inference-time strong generalization to handle complicated, unfamiliar codebases.
We identify that the prevailing Task-Centric Learning (TCL) approach, which scales exposure across disparate repositories, fails to address this limitation.
In response, we propose Repository-Centric Learning (RCL), a shift that prioritizes vertical, single-repository depth over horizontal, single-task breadth for SLM training.
To accomplish this depth, we design a four-unit Repository-Centric Experience, transforming static codebases into interactive learning signals,
to train \model-4B, a family of highly compact models built as repo-specialized experts. Under controlled comparisons with the same teacher and data budget, RCL consistently outperforms TCL across tasks, teachers, model scales, and programming languages. On seven target repositories, \model-4B experts match or exceed existing open-weight models up to $8\times$ larger (e.g., CWM by Meta, Qwen3-Coder-30B) and surpass or match efficiency-focused commercial models (e.g., GPT-4.1-mini, GPT-5-nano) across multiple SWE tasks.
Further analysis reveals that RCL yields higher training and inference efficiency, deep knowledge acquisition, and
can scale to repository-cluster experts.
Our work emphasizes that for building efficient intelligence,
repository mastery is a distinct, necessary, and practical dimension that complements general coding capability.

  
\end{abstract}
\section{Introduction}
\label{sec:intro}

Large language model (LLM) agents have significantly improved productivity across software engineering workflows. Yet today's most widely adopted systems~\cite{claude_code, gemini_cli, cursor, copilot, kiro} depend on proprietary, closed-source models with hundreds of billions of parameters~\cite{claude4-5, gemini-3-pro, gpt-5}, making them costly to operate and unsuitable for local deployment in privacy-sensitive settings. As agentic workflows transition from experimental use to production, there is a growing demand for capable \emph{small language models} (SLMs) with fewer than 10B parameters that can be deployed locally and cost-effectively~\cite{li2025appleintelligencefoundationlanguage}. SLMs critically preserve code confidentiality while avoiding dependence on proprietary APIs and their evolving economic and access constraints~\cite{cursor-2025-composer}.

\paragraph{The current approach}
To narrow the capability gap between small open-weight models and much larger proprietary systems, recent work has largely centered around \textbf{Task-Centric Learning (TCL)}, predominately framing software engineering as GitHub issue resolution~\cite{jimenez2024swebench}. Early work relied on fixed-workflow scaffolds~\cite{xia2024agentless, li2025patchpilot}, decomposing issue resolution into stages including retrieval, localization, patch generation, and validation, and training specialized models for each subtask~\cite{tang2025copatcher, xie-2025-swefixer}. More recent work, driven by Reinforcement Learning with Verifiable Rewards~\cite{guo2025deepseek}, replaces fixed pipelines with flexible agent-driven exploration, using the correctness of the final patch as supervision~\cite{wei2025swerl}. To scale this approach, several ``gym'' infrastructures have been introduced to support large-scale TCL training across diverse, unrelated repositories~\cite{deepswe2025, SWESwiss2025, zan2025multiswebenchmultilingualbenchmarkissue, deng2025swebenchproaiagents}. The shared assumption is that exposure to massive numbers of task trajectories will yield broadly generalizable bug-fixing skills, producing increasingly capable open-weight \textit{issue resolvers}. (See \cref{sec:related} for more related work.)

\paragraph{The limitation}
While TCL has shown promise for medium-to-large models (e.g., 30B+ parameters), it exposes fundamental limitations when applied to SLMs. The core issue is an inference-time capability gap: frontier large models can compensate for missing prior knowledge in unfamiliar environments through extensive inference-time search, trial-and-
error, and reflection.
In contrast, SLMs lack this capacity: When trained as task specialists across disparate codebases, they tend to overfit to surface-level patterns (e.g., patch templates, shell syntax)~\cite{wang-etal-2023-recode} rather than internalizing repository-specific design and semantics. Without grounded understanding of the target codebase, SLMs struggle to navigate through and adapt to unseen environments, leading to inefficient search and brittle generalization at deployment, as we demonstrate in \cref{sec:eval}.

\paragraph{Our proposal}
These observations motivate a shift in perspective:
If SLMs cannot reliably acquire repo-specific knowledge at inference time, that knowledge must be internalized \emph{during training} and carried into deployment as a prior.
We formalize this as \textbf{Repository-Centric Learning (RCL)}: while TCL holds the task fixed (issue resolution) and varies the repository, RCL holds the repository fixed and varies the task across the full software engineering lifecycle. This inversion of the training axis mirrors how human developers build expertise---not by fixing isolated bugs across thousands of unrelated projects, but through sustained, multi-modal interaction with a target codebase that includes reading, testing, debugging, and iterative modification. Over time, these experiences induce an implicit representation of the repository's \emph{operational semantics}: its dependency structure, execution dynamics, and behavioral invariants. To our knowledge, we provide the first systematic operationalization of repo-specific multi-task training for the SWE agent community, and show that for SLMs it is consistently superior to scaling breadth.

We emphasize that RCL complements rather than replaces task-centric learning (TCL).
While TCL optimizes a \emph{single task}, typically issue resolution, across many unrelated repositories to train a general-purpose issue resolver, RCL targets \emph{multiple software engineering tasks} within a \emph{fixed repository}, yielding a repo-specialized model.
For a given repository, TCL may resolve individual issues in isolation, whereas RCL enables consistent performance across tasks within the same codebase, including issue resolution, feature implementation, test development, etc.
This distinction reflects a key requirement of repository-level SWE performance: combining task-agnostic capability with repository-specific knowledge. While large models may recover the latter at inference time through strong generalization, RCL is necessary for SLMs to internalize this knowledge during training.
RCL thus deliberately trades cross-repository generality for depth, targeting settings where the repositories of interest are known in advance, such as an organization's private codebases.

To operationalize RCL, we train SLMs not merely as issue resolvers, but as \emph{repository experts}, learning from diverse signals that span the full software engineering lifecycle of a target codebase. We structure this training around four units of \textbf{Repository-Centric Experience (RCX)} (\cref{fig:overview}):
(i) \textit{Software Design}, where the agent infers architectural intent and module responsibilities through \emph{active} code analysis;
(ii) \textit{Contextual Implementation}, where it models cross-file dependencies, internal APIs, and implementation logic via context-aware completion;
(iii) \textit{Evolutionary Replay}, where it learns historical changes to understand design trade-offs and constraints shaping the codebase over time;
(iv) \textit{Semantic-Runtime Alignment}, where it designs tests to capture nuanced discrepancies between specifications and runtime behaviors.
Together, these experiences enable SLMs to acquire repo-specific knowledge that supports robust, multi-task performance within a single codebase.


\paragraph{Results}
We validate RCL by training \model, a family of \emph{only 4B-parameter} models specialized
as \underline{S}mall Re\underline{\smash{po}}-Exper\underline{t}s in target codebases.
Under a \textbf{repository-centric evaluation} with a strict temporal protocol which spans four software engineering tasks (issue resolution, feature implementation, test generation, and codebase QA), \model consistently breaks the scaling trends established by TCL. In controlled comparisons where both share the same teacher, student, and data budget, RCL outperforms TCL on every task, and this advantage also holds when we switch to a weaker teacher, scale the student up to 30B, or move to a JavaScript repository. Despite its compact size, \model matches or exceeds open-weight TCL-trained models up to 
8x larger on our seven target repositories, and matches or exceeds efficiency-comparable commercial models such as GPT-5-nano, GPT-4.1-mini, and Gemini-2.5-Flash-Lite.

Beyond end-to-end accuracy, RCL induces a qualitative shift in agent behavior. By internalizing repository-specific structure and dynamics, \model achieves \emph{higher training sample efficiency} and \emph{lower inference-time cost}.
Crucially, RCL remains necessary even in the combination with strong context retrieval, establishing a clear performance gap that indicates \textit{deep parametric knowledge acquisition} beyond simple context memorization or superficial pattern matching.
Ablation studies further show that the proposed units of RCX are synergistic: for example, training on test generation improves issue resolution, confirming that repository mastery emerges from holistic exposure rather than isolated tasks.

Our contributions are:

\begin{itemize}[nosep, leftmargin=*]
\item \textbf{The Paradigm:}
We introduce \emph{Repository-Centric Learning (RCL)} as a new axis for training coding agents. We show that particularly for SLMs, task breadth alone is insufficient: repo-specific knowledge must be internalized during training to compensate limited inference-time capacity.


\item \textbf{The Framework:}
We present a practical RCL framework that converts static repositories into interactive learning signals via four-unit \textit{Repository-Centric Experiences},
mirroring how human experts acquire codebase mastery.


\item \textbf{The Results:}
We train \model-4B using RCL and demonstrate a clear break from task-centric scaling trends. Under controlled comparisons, RCL outperforms TCL across tasks, teachers, model scales, and languages; on seven target repositories, \model-4B matches or exceeds much larger open-weight models (30B+) and efficiency-comparable commercial APIs.


\item \textbf{The Analysis:} We show that RCL yields higher sample efficiency, lower inference cost, and multi-task synergy. Ablations confirm that these gains persist beyond context retrieval and require sufficient learning capacity, indicating deep parametric knowledge acquisition rather than surface-level memorization.
We also use linear probing to predict inter-repository synergy or interference, and thus unlocks predictable cross-repository learning to build repo-cluster-experts.
We also validate RCL's practicality under repo-evolution and sparse commit history.

\end{itemize}


\section{Repository-Centric Learning}
\label{sec:method}

\begin{figure*}[t]
    \centering
    \includegraphics[width=0.99\textwidth]{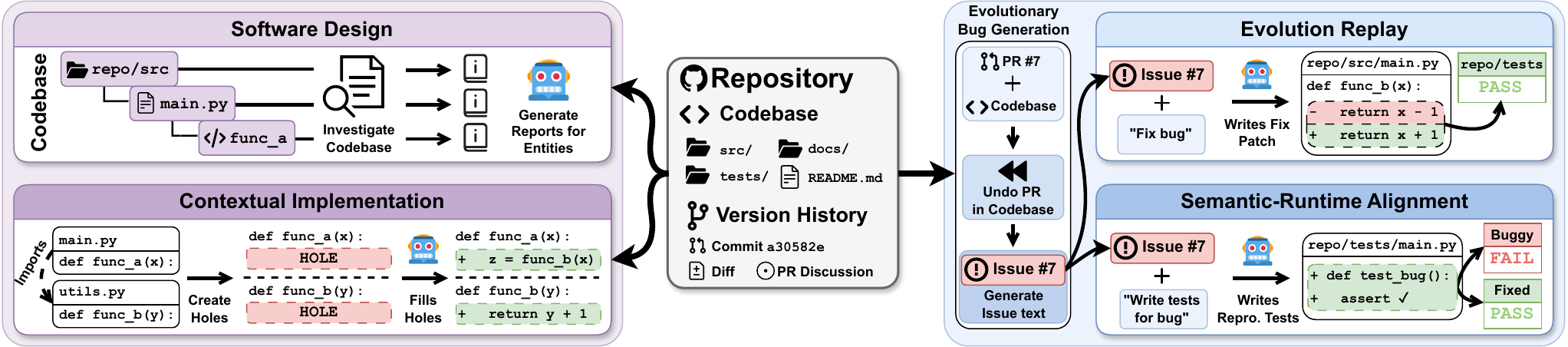}
    \caption{
    {A Four-Unit Design of \textbf{Repository-Centric Experience (RCX)}}
    }
    \label{fig:overview}
\end{figure*}

As discussed in \cref{sec:intro}, Task-Centric Learning (TCL) builds broad task skills but leaves small models without the repository-specific grounding needed to work effectively in a target codebase.
Classical domain adaptation through continued pre-training on static source files, including structural variants such as fill-in-the-middle (FIM), is also insufficient for agentic software engineering: it teaches the model to recite code more than to navigate, test, debug, and modify a live repository.
We therefore propose \textbf{Repository-Centric Learning (RCL)}, which trains a model from diverse \textbf{Repository-Centric EXperience (RCX)} collected inside the target repository.
Each RCX is a learnable agentic trajectory in which the agent observes repository state, searches for context, reasons about code behavior, and produces edits, tests, or reports.
This turns repository knowledge into interactive supervision, allowing the model to internalize the repository's ``physics'' before deployment.

To cover the main forms of repository expertise, we organize RCX into four units, shown in \cref{fig:overview}.
Together, they expose the model to design intent, implementation structure, evolution history, and runtime semantics.
Full construction details are provided in \cref{app:RCX}.


\paragraph{Software Design}
Static code encodes architectural intent, design philosophy, and cross-file dependencies that are only partly captured by syntactic analysis.
We extract this semantic layer through \textit{active code analysis}, inspired by \citet{lin2025learningfactsscaleactive}: given a target module, file, function, or class, an agent explores the repository and reports the component's functionality, design rationale, and interactions with the broader system.
This teaches the model why the code is organized as it is, not only what each fragment does.

\paragraph{Contextual Implementation}
Implementing functionality that complies with an existing codebase requires awareness of global conventions, abstractions, and cross-file dependencies.
We adapt FIM into \textit{agentic FIM}: given missing functionality, the agent must actively explore the repository to discover relevant context before producing a compliant implementation.
This encourages the model to internalize repository structure rather than relying on externally supplied retrieval context.

\paragraph{Evolutionary Replay}
A core maintainer's expertise emerges not from reading one version of the codebase alone, but from experiencing the repository's evolution over time.
We replicate this signal by mining pull request history to reintroduce historical bugs into the current codebase, then tasking the agent with resolving them.
We adapt PR Mirror~\cite{yang2025swesmith} for repository-centric coverage and augment it with synthetic tasks when commit history is sparse.
Debugging these faults teaches the model not only what the code does, but how it has failed and been repaired over time.


\paragraph{Semantic-Runtime Alignment}
Bugs arise from misalignment between expected program behavior and actual runtime execution.
We train the agent to formalize such misalignments by reusing the Evolutionary Replay instances for test generation: instead of fixing the bug, the agent writes \textit{reproduction tests} that fail on the buggy code and pass on the fixed codebase.
This extracts an additional learning signal from the same repository history and teaches the model to connect natural-language specifications with executable behavior.


In this work, we instantiate RCL by synthesizing RCX trajectories with a teacher model (Gemini-2.5-Pro) and training a student SLM (Qwen3-4B-Instruct-2507) via supervised fine-tuning (SFT).
This implementation is sufficient to demonstrate the benefits and necessity of RCL in the following sections, but RCL idea generally is not tied to SFT or to this particular teacher-student recipe.

\section{Experimental Setup}
\label{sec:setup}


\subsection{Repository-Centric Evaluation for Agentic Coding}
To rigorously assess the efficacy of RCL, we design a comprehensive \textbf{Repository-Centric Evaluation (RCE)} suite.
Unlike traditional Task-Centric Evaluation (TCE), which focuses on scaling the repository diversity within a single task to test generalization to wild repositories,
we propose RCE to validate whether RCL enables multi-spectrum mastery in the target codebase rather than only fitting to specific task formats (e.g., issue resolving).

Specifically, we compile several existing TCE benchmarks across four agentic coding tasks:

\begin{itemize}[nosep, leftmargin=*]
    \item \textbf{Issue Resolution:}
    We employ SWE-bench-Verified \cite{Introduc24:online} to assess the agent's ability to resolve realistic GitHub issues.
    
    \item \textbf{Test Generation:}
    Derived from SWE-bench-Verified, TDD-Bench-Verified \cite{otter} evaluates the capability of writing reproduction tests, i.e., test cases that fail on a buggy codebase but pass on the patched version.

    \item \textbf{Feature Implementation:}
    FEA-Bench \cite{feabench} selects realistic Pull Requests with a focus on new feature development.
    It can be quite challenging as it often requires scattered cross-file modifications with compliance: its gold patches touch 3.42 files and change 140.7 lines on average (vs.\ 1.19 files and 13.3 lines in SWE-bench-Verified), and the best model evaluated by \citet{feabench} resolves only about 14\%.
    
    \item \textbf{Codebase QA:}
    Separate from explicit coding tasks, SWE-QA \cite{peng2025swe} evaluates the ability to answer codebase-specific questions, a practical skill in human-AI collaborative development.
\end{itemize}

\paragraph{Temporal evaluation and repository selection}
A critical challenge in evaluating repository experts is data leakage.
To ensure  the model is not evaluated on changes that it has already learned during RCL,
we implement a strict \textbf{Temporal Evaluation Protocol} with a knowledge-cutoff date of December 31, 2020.
We construct our training data using only repository knowledge available prior to this date, and utilize instances created after this cutoff as our RCE instances.
To manage computational costs without loss generality, we select the top-7 repositories with the highest density of post-cutoff instances across the selected four benchmarks above.
Our final RCE suite consists of 231 issue resolving, 211 test generation, 152 feature implementation, and 336 codebase QA instances.

\paragraph{Metrics}
For the three coding tasks (issue resolution, test generation, feature implementation), we report the pass rate (pass@1) based on execution results.
For codebase QA, following the protocol by SWE-QA \cite{peng2025swe}, we utilize LLM-as-a-judge (based on GPT-5.2) to score responses by agents, and report the average scaled to 0-100.



    
    

\subsection{Implementation Details}


\paragraph{Data}
To ensure fair comparisons between RCL and TCL, by default, we always follow a fixed teacher-student distillation setup: Gemini-2.5-Pro as the teacher, Qwen3-4B-Instruct-2507 as the student (base model).
For TCL, we gather wild repositories and their problem instances other than the target evaluation repositories, such as those from SWE-Smith \cite{yang2025swesmith}, and use the teacher model to synthesize trajectories by working on problem instances of a fixed task.
For RCL, we let the teacher model to work within a snapshot of the target repository that was released \emph{before} the knowledge cutoff date to obtain training data.
We make sure TCL and RCL match the volume of training samples.
Off-the-shelf models (\cref{tab:rq1}) are the only exception.

\paragraph{Inference}
To ensure a fair comparison focused on model capability rather than agentic scaffold engineering, we utilize mini-SWE-agent, which is designed for \textit{model-centric evaluation} \cite{mini-swe-agent2026} across all evaluations.
We make minimal changes to adapt the default prompt to the four evaluation tasks to ensure models can fully understand the task requirements.
For all open-weight models including our trained ones, we apply the following sampling parameters adjusted from the recommendation of the base model \cite{qwen3technicalreport}: a temperature of $1.0$, top-$p$ of $0.8$, top-$k$ of $20$, and a repetition penalty of $1.05$.
For commercial LLM APIs, we keep the default settings and disable or minimize the reasoning effort to focus on non-reasoning models in our evaluation.
To finish the costly long-trajectory evaluation within a reasonable time, we set the maximum trajectory steps to 64 and the context limit to 48k tokens. Trial runs confirmed that this configuration helps prevent getting stuck in super-long trajectories without degrading most normally advancing trajectories.
Since we do not use greedy decoding, all metrics are produced by 3 independent inference runs to mainly report the average, and unless otherwise noted, error bars resulting from this in all figures are of 1-sigma.

\paragraph{Training}
We perform Supervised Fine-Tuning (SFT) using the ms-swift library \cite{zhao2024swiftascalablelightweightinfrastructure} with Megatron \cite{megatron-lm} on 2$\times$H200 GPUs.
By default, we train 2 epochs for all experiments with a global batch size of 16 and a maximum sequence length of 32,768 tokens.
We use AdamW optimizer with learning rate of $1e-5$, cosine decay to $1e-6$, and linear warmup fraction of 0.03.
For a typical 8k-sample 4-unit RCL, it takes about 5 hours.

\section{Evaluation and Results}
\label{sec:eval}

\subsection{RCL versus TCL under Controlled Conditions}
\label{sec:rq1}

We first compare RCL against TCL under a controlled setup, where both share the same teacher, student, and training data volume and differ only in whether the training data comes from the target repository (\cref{sec:setup}). This directly isolates the effect of RCL and constitutes our main evidence.

\begin{table}[t]
\centering
\caption{\textbf{Controlled Comparison between RCL and TCL.} Bottom rows: $^{\dagger}$weaker teacher, $^{\ddagger}$30B student, $^{\S}$JavaScript repository.}
\label{tab:singletask}
\small
\begin{tabular}{lccc}
\toprule
\textbf{Task} & \textbf{Target Repo} & \textbf{TCL} & \textbf{RCL} \\
\midrule
Issue Resolving & Django & 16.95 & \textbf{20.62} \\
Test Generation & Django & 4.09 & \textbf{11.01} \\
Feature Impl. & Sympy & 2.15 & \textbf{4.84} \\
Codebase QA & Django & 71.59 & \textbf{75.79} \\
\midrule
Issue Res.$^{\dagger}$ & Django & 17.3 & \textbf{21.7} \\
Exec Avg$^{\ddagger}$ & Django & 20.1 & \textbf{34.9} \\
Issue Res.$^{\S}$ & Preact (JS) & 5.9 & \textbf{21.6} \\
\bottomrule
\end{tabular}
\end{table}

\paragraph{RCL advantage persists across SWE tasks}
Comparing RCL against TCL with the same training data volume (2k) by training and evaluating on each task respectively, \cref{tab:singletask} confirms the advantage of RCL is not restricted to one certain task, and therefore applicable to improve various coding tasks. These results also isolate RCL's advantage from the effect of multi-task learning.


\paragraph{RCL advantage remains with weaker teacher, larger student model, and different language}
We further repeat the controlled comparison while changing one factor of the default setup at a time (\cref{tab:singletask}, bottom), keeping the student, teacher, and data volume of RCL and TCL identical within each comparison.
\textit{Teacher:} with the weaker Gemini-2.5-Flash as teacher, RCL still outperforms TCL on Django issue resolution (21.7\% vs.\ 17.3\%), indicating the gain comes from the RCX structure rather than purely the teacher quality.
\textit{Student model scale:} with Qwen3-Coder-30B as the student, $8\times$ larger than our default 4B student, RCL reaches an Exec Avg of 34.9\% versus 20.1\% for TCL, so RCL also benefits larger student models to some extent.
\textit{Language:} we target the JavaScript repository preact. TCL learns from 4k teacher trajectories over problem instances from 74 JavaScript repositories in SWE-Smith Multilingual, while RCL learns from 4k RCX trajectories synthesized inside a preact snapshot that predates its earliest evaluation instance, preserving our temporal protocol. On all 17 preact instances in SWE-bench Multilingual, RCL resolves 21.6\% of issues versus 5.9\% for TCL, so RCL is not specific to Python.
We choose JavaScript because it is the only other language supported by SWE-Smith Multilingual, which is needed to build the TCL baseline, and preact has the most JavaScript instances in SWE-bench Multilingual.

\begin{figure*}[t]
    \centering
    \begin{subfigure}[t]{0.37\textwidth}
        \centering
        \includegraphics[width=\linewidth]{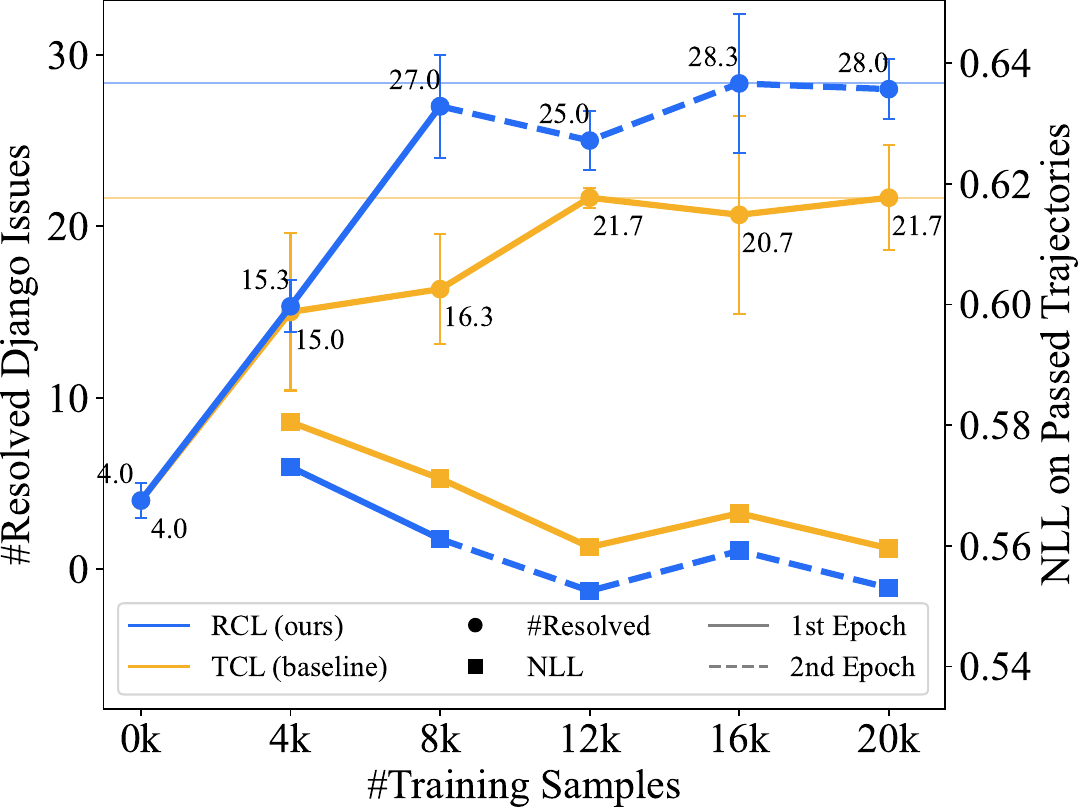}
        \caption{}
        \label{fig:scaling_curve}
    \end{subfigure}
    \hfill
    \begin{subfigure}[t]{0.26\textwidth}
        \centering
        \includegraphics[width=\linewidth]{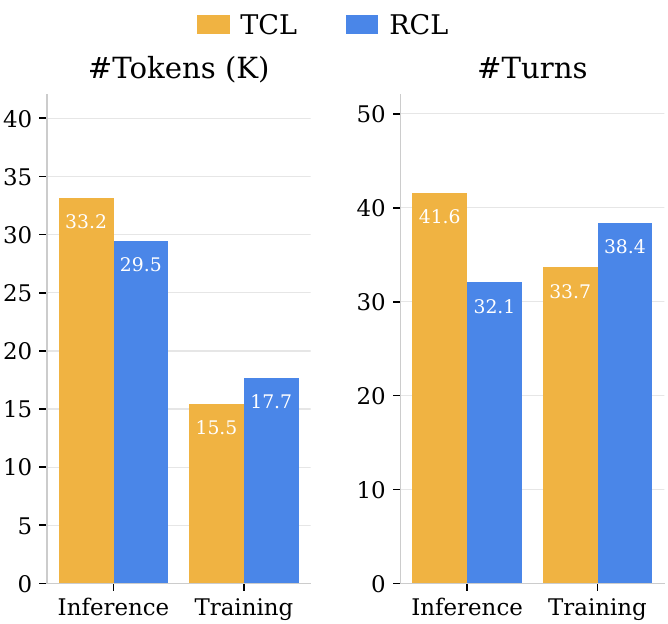}
        \caption{}
        \label{fig:efficiency}
    \end{subfigure}
    \hfill
    \begin{subfigure}[t]{0.31\textwidth}
        \centering
        \includegraphics[width=\linewidth]{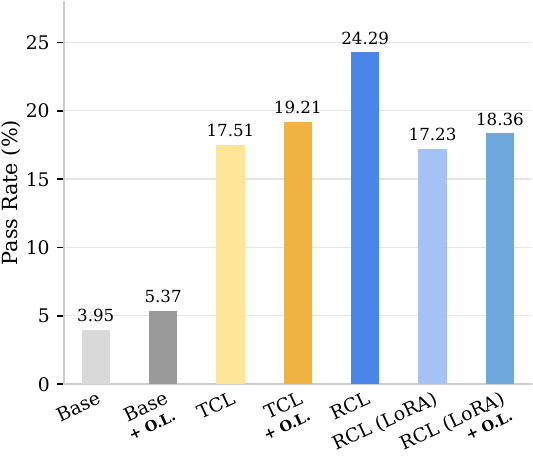}
        \caption{}
        \label{fig:nature}
    \end{subfigure}
    \caption{
    \textbf{(a) Learning Sample Efficiency} (RCL: 10k Django samples, 2 epochs; TCL: 20k SWE-Smith samples, 1 epoch).
    \textbf{(b) Inference Cost.}
    \textbf{(c) Comparison of Optimization Strategies} on Django issue resolution. \textbf{O.L.}: oracle localization; \textbf{LoRA}: rank $r=128$.
    }
    \label{fig:combined_efficiency}
\end{figure*}

\paragraph{Higher training and inference efficiency}
By evaluating intermediate checkpoints during training, \cref{fig:scaling_curve} shows the scaling trends with respect to the amount of training samples: RCL surpasses the peak performance of TCL using approximately 2.5$\times$ fewer training budget.
Meanwhile, by measuring the length distributions of both training and inference trajectories, \cref{fig:efficiency} shows that RCL leads to shorter (11.1\% less total token consumption and 23\% fewer agentic turns) inference trajectories despite being trained with longer (12.6\% more tokens) ones.
The higher sample efficiency and lower inference cost further demonstrate the advantage of RCL over TCL.

\begin{table}[t]
\centering
\caption{\textbf{Leave-one-out Ablation of RCX Units} (Django).}
\label{tab:leaveoneout}
\resizebox{\columnwidth}{!}{%
\begin{tabular}{lcccc}
\toprule
\textbf{Configuration} & \textbf{Issue Res.} & \textbf{Test Gen.} & \textbf{Feat. Impl.} & \textbf{Code QA} \\
\midrule
\textbf{Full-unit RCL} & \textbf{24.01} & \textbf{16.98} & \textbf{14.29} & \textbf{83.11} \\
\midrule
\quad $-$ Evolutionary Replay & 16.10 & 15.72 & 14.29 & 78.58 \\
\quad $-$ Sem.-Rt. Alignment & 23.45 & 3.14 & 9.52 & 80.60 \\
\quad $-$ Contextual Impl. & 20.62 & 16.38 & 4.76 & 79.44 \\
\quad $-$ Software Design & 22.32 & 14.15 & 14.29 & 75.77 \\
\bottomrule
\end{tabular}%
}
\end{table}

\paragraph{Multi-task RCL further improves performance}
To validate the contribution of each of the four RCX unit in our design, we conduct a leave-one-out study with the total training data volume fixed at 6k samples. Results in \cref{tab:leaveoneout} show that each unit not only effectively contributes to the performance of its corresponding task, but also often improves other tasks, indicating cross-task synergy in repo-specific learning.

\subsection{Comparison with Off-the-Shelf Models}
\label{sec:offshelf}

To show that RCL produces practically useful agents, we further compare \model-4B, trained with 8k four-unit RCL samples per target repository, against off-the-shelf models in \cref{tab:rq1}, including TCL-trained coding agents that learn from orders of magnitude more data. Our results show that repository experts trained with RCL match or outperform much larger generalist models on the repositories they are built for.

\begin{table*}[t]
\centering
\caption{
\textbf{
End-to-end Repository-Centric Evaluation on Four Agentic SWE Tasks.}
\textbf{Exec Avg}: average pass rate over the three execution-based tasks.
}
\label{tab:rq1}
\resizebox{0.97\textwidth}{!}{%
\begin{tabular}{lccccccc}
\toprule

\multirow{2}{*}{\textbf{Model}} & \multirow{2}{*}{\textbf{Size}} & \multirow{2}{*}{\textbf{SWE Training Data}} & \textbf{Issue Resolving} & \textbf{Test Gen} & \textbf{Feature Impl} & \textbf{Exec Avg} & \textbf{Codebase QA} \\
& & \textbf{} & (Pass \%) & (Pass \%) & (Pass \%) & (Pass \%) & (Score 0-100) \\

\midrule

\multicolumn{8}{l}{\textit{\textbf{Efficency-focused Commercial LLM APIs}}} \\
GPT-4.1-mini & [10, 100]B & - & \underline{21.79} & \underline{22.27} & \underline{5.70} & \textbf{17.85} & \textbf{80.28} \\
Gemini-2.5-Flash-Lite & [0.1, 10]B & - & 8.80 & 10.74 & 0.88 & 7.46 & 73.36 \\
GPT-5-nano & [0.1, 10]B & - & 6.49 & 0.00 & 1.75 & 2.97 & 35.64 \\

\cmidrule(lr){1-8}

\multicolumn{8}{l}{\textit{\textbf{Open-Weight Generalist Models}}} \\
Gemma-3-27b-it & 27B & - & 1.29 & 2.37 & 0.00 & 1.34 & 65.19 \\
Qwen3-Coder-30B & 32B & - & 16.74 & 11.85 & 3.29 & 11.56 & 65.48 \\
Qwen3-4B-Instruct-2507 & 4B & - & 3.32 & 5.37 & 0.00 & 3.20 & 61.79 \\

\cmidrule(lr){1-8}

\multicolumn{8}{l}{\textit{\textbf{Coding Agents Trained with TCL}}} \\
CWM (Meta) & 32B & 3M trajs from 3.15K repos & \textbf{22.22} & 17.38 & 4.17 & 15.88 & 73.09 \\
Mini-Coder-4B & 4B & 400K trajs from 128 repos & 18.76 & 0.63 & 4.61 & 8.70 & 57.30 \\

\cmidrule(lr){1-8}

\multicolumn{8}{l}{\textit{\textbf{RCL (ours)}}} \\
{\model} & \textbf{4B} & \textbf{8K} trajs in \textbf{one} target repo & 19.34 & \textbf{22.75} & \textbf{5.92} & \underline{17.12} & \underline{78.05} \\

\bottomrule
\end{tabular}%
}
\end{table*}

\paragraph{Better overall performance}
\cref{tab:rq1} shows that our RCL-trained model, \model-4B, breaks the scaling barriers observed in task-centric learning.
Despite having only 4B parameters and being trained with much less trajectories, \model-4B achieves an average execution-based score of 17.12\%, matching or exceeding open-weight models up to 8× larger,
including {Qwen3-Coder-30B-A3B-Instruct}, {Gemma-3-27B-it}, and {CWM-32B}.
Paired permutation tests over all 594 execution-based instances (\cref{tab:significance}) confirm that it significantly outperforms Qwen3-Coder-30B and Mini-Coder-4B, and statistically matches CWM-32B and GPT-4.1-mini on Exec Avg.
Notably, \model-4B also surpasses or effectively matches efficiency-comparable commercial APIs such as {GPT-4.1-mini}, {GPT-5-nano}, and {Gemini-2.5-Flash-Lite}.
This advantage also extends beyond one single task (issue resolution). Though TCL-trained models (Mini-Coder-4B, CWM-32B) achieve comparable performance on issue resolution by training on much more data, they fail to generalize well to other realistic tasks and fall behind \model-4B by an obvious margin, confirming our recipe can better meet practical multi-task demands.

\paragraph{Statistical significance}
For each instance, we compute the pass probability as the mean over the three inference runs, and test the mean paired difference $d = p(\text{\model}) - p(\text{baseline})$ with a two-sided paired sign-flip permutation test (100k permutations).
We report 95\% paired-bootstrap confidence intervals and apply Holm--Bonferroni correction.
All models are evaluated on identical instance sets with the same mini-SWE-agent harness, so the tests are fully paired.
\Cref{tab:significance} reports the results over all 594 execution-based instances and for codebase QA.

\begin{table}[t]
\centering
\caption{\textbf{Paired significance tests} for \model-4B against off-the-shelf baselines. Gap = \model $-$ baseline (percentage points for Exec Avg, score points for QA).}
\label{tab:significance}
\resizebox{\columnwidth}{!}{%
\begin{tabular}{llccc}
\toprule
\textbf{Metric} & \textbf{Baseline} & \textbf{Gap} & \textbf{95\% CI} & \textbf{$p$ (Holm)} \\
\midrule
\multirow{4}{*}{Exec Avg} & CWM-32B & +0.90 & [$-$1.35, +3.17] & 0.903 \\
 & Qwen3-Coder-30B & +5.89 & [+3.31, +8.47] & 6.0e-5 \\
 & Mini-Coder-4B & +8.19 & [+5.95, +10.49] & 4.0e-5 \\
 & GPT-4.1-mini & $-$0.31 & [$-$2.86, +2.19] & 0.903 \\
\midrule
\multirow{4}{*}{Codebase QA} & CWM-32B & +4.97 & [+3.58, +6.33] & 1.6e-4 \\
 & Qwen3-Coder-30B & +12.57 & [+10.34, +14.84] & 1.6e-4 \\
 & Mini-Coder-4B & +20.75 & [+18.55, +23.06] & 1.6e-4 \\
 & GPT-4.1-mini & $-$4.60 & [$-$5.75, $-$3.47] & 1.6e-4 \\
\bottomrule
\end{tabular}%
}
\end{table}

\subsection{The Necessity and Nature of RCL}
\label{sec:rq2}

Having demonstrated the performance and efficiency advantages of RCL, we now investigate the possible reasons behind these improvements.
Specifically, we question
whether RCL simply acts as a proxy for better retrieval,
and whether the acquired capability is shallow (style alignment) or deep (parametric knowledge).

\paragraph{RCL is irreplaceable by context retrieval}
A natural counter-hypothesis to our approach is that RCL's advantage comes from memorizing file locations, acting as an implicit retriever to alleviate the burden of codebase exploration. If this were true, a TCL model equipped with perfect contextual knowledge should match RCL.
We test this by providing "Oracle Localization"---injecting the locations of ground-truth patched files and functions as hints into the prompt---thereby removing the need of navigation and proxying the presence of an excellent retrieval system.
As shown in \cref{fig:nature}, oracle localization indeed improves the base model (3.95\% $\rightarrow$ 5.37\%) and TCL model (17.51\% $\rightarrow$ 19.21\%), but they still underperform the RCL model without any hints (24.29\%) by a large gap,
indicating RCL goes beyond memorization via parametric knowledge injection.

\paragraph{RCL requires sufficient learning capacity}
We further probe the nature of RCL by comparing full parameter fine-tuning v.s. LoRA, with all other RCL setups fixed.
Because if LoRA performs competitively, then it indicates RCL primarily adapts the model to the repository's superficial coding styles or to the agentic scaffold's response format, achievable with much lower learning capacity than deep repo-specific knowledge acquisition.
However, as shown in \cref{fig:nature}, the LoRA-tuned RCL model (even with a high rank 128, updating $\approx$5.38\% of full parameters) significantly lags behind the fully fine-tuned model (17.23\% v.s. 24.29\%), dropping to a performance level similar to the TCL model.
Notably, this limitation persists even when perfect localization hint is provided (18.36\%), confirming that the performance gap is not due to navigational errors but a fundamental lack of capacity.
This further implies that RCL involves substantial knowledge and dynamics internalization, thereby requiring sufficient learning capacity of full-weight updates.

\subsection{Predictable Inter-Repository Transfer Enables Cluster-level RCL}
\label{sec:transfer}

Given the demonstrated advantage of RCL, it is still unclear what blocks the learning across wild repositories (TCL) from producing the optimal performance on a target repository.
Specifically, when does learning from a certain repository benefit or hurt the performance on other ones?
In this section, we show
(1) inter-repo synergy (positive transfer) and interference (negative transfer) both exist, (2) transfer effect is predictable by probing repository likeness from base model activations, and (3) jointly learning from a cluster of similar repositories can outperform single-repository learning.


\begin{figure*}[t]
    \centering
    \hfill
    \begin{subfigure}[t]{0.37\linewidth}
        \centering
        \includegraphics[width=\linewidth]{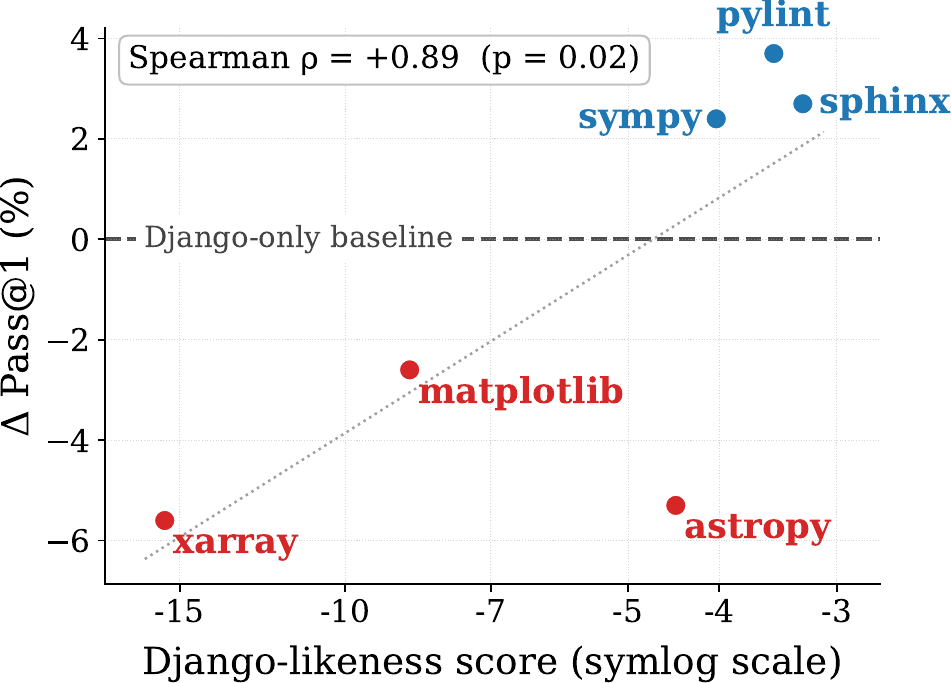}
        \caption{}
        \label{fig:pair_transfer}
    \end{subfigure}
    \hfill
    \begin{subfigure}[t]{0.48\linewidth}
        \centering
        \includegraphics[width=\linewidth]{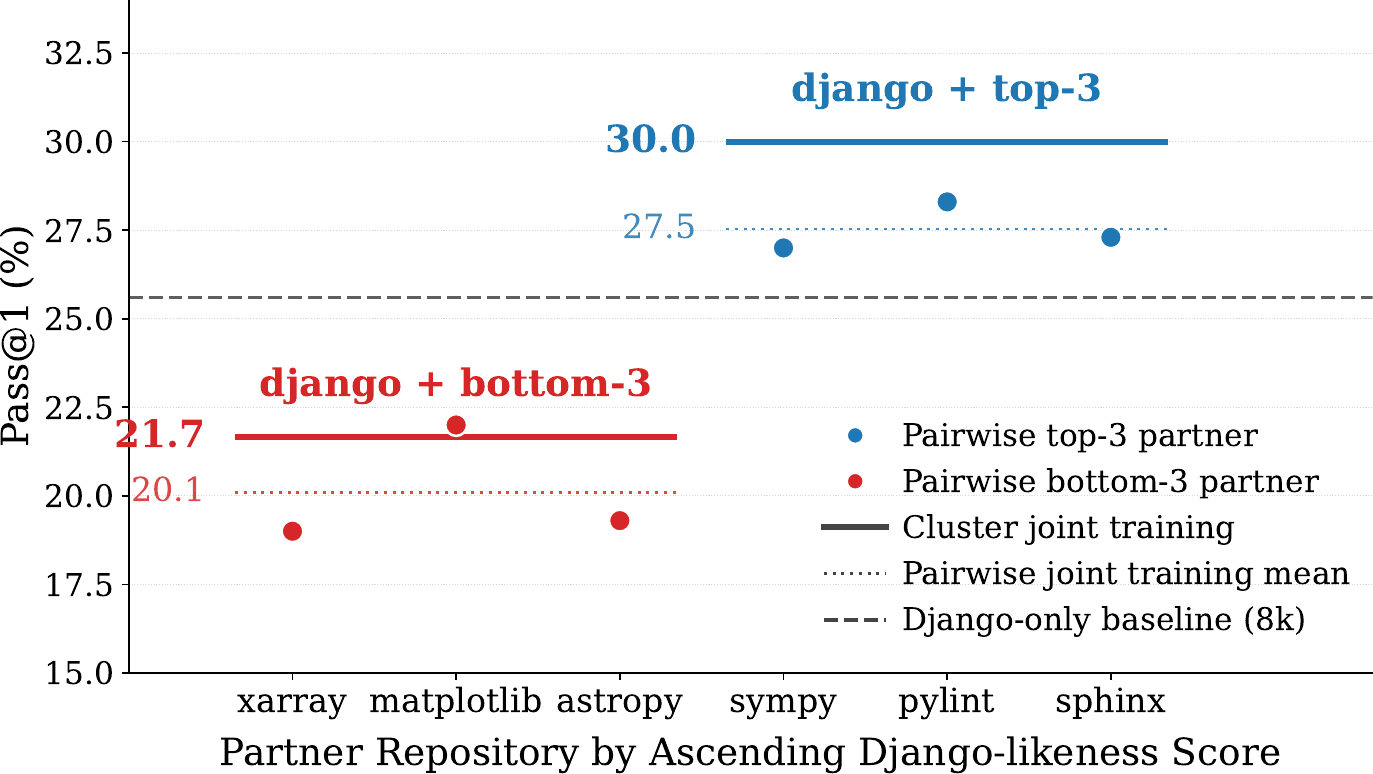}
        \caption{}
        \label{fig:cluster_transfer}
    \end{subfigure}
    \hfill
    \caption{
    \textbf{Probe-predicted repository likeness explains and guides cross-repository transfer}.
    (a) Pairwise transfer effect on Django vs.\ probe likeness score. (b) Joint training on the positive (Django + top-3) vs.\ negative (Django + bottom-3) cluster.
    }
    \label{fig:transfer_combined}
\end{figure*}

\paragraph{Transfer effect varies in sign and magnitude across partners}
Fixing Django as the target repo,
we measure the effect of jointly learning another repository on Django's issue resolution performance.
For example, when the data budget is matched at 4k, compared to learning from Django alone, we evaluate whether learning from Django \emph{and} Sympy improve or degrade the performance on Django.
We train and evaluate six such (Django, X) pairs using our selected seven repositories.
As we can see in \cref{fig:pair_transfer} (comparing the Y-axis values),
the transfer effect ranges from +3.7\% (pylint) to -5.6\% (xarray),
with three pairs producing positive transfer and the other three producing negative ones,
revealing the existence of both inter-repository synergy and interference.


\paragraph{A linear probe on base-model activations predicts transfer effects}
Embedding models offer a tempting approach to explain positive and negative transfer by measuring semantic distances in embedding space, but our pilot study with gemini-embedding-2 yielded no meaningful signal with very close distances between different repository-pairs. This suggests a possible inability to represent and distinguish deeper semantic-level features.
Instead, we train a linear probe (L2-regularized logistic
regression) to discriminate Django code from non-Django code using last-token activations from our base model, inspired by the representation engineering literature \cite{alain2018understandingintermediatelayersusing, zou2025representationengineeringtopdownapproach}.
During the probe training, we feed a code chunk (valid AST span) either from Django or another repository into the base model to produce last-token activations from the last layer's residual stream, which the probe exclusively uses to predict whether the input code is from Django or not.
As such, the probe learns to identify the student model's representation of the Django code in its activation space, and an AUC of 0.998 on a held-out set confirms the successful learning.

We define the target-repo-likeness (Django-likeness here) score as the probe's output logit value for believing the input is from the target repo, and measure its correlation with pairwise inter-repository transfer delta in \cref{fig:pair_transfer}:
the three repositories that have top-3 likeness scores are exactly those that induce positive transfer, and the bottom-3 with lowest likeness scores are those that induce negative transfer, showing that the probe successfully predicts transfer effects.


\paragraph{Probe-selected repo-clusters yield principled multi-repository learning}
To validate that the probe's predictions are actionable for the top-3 and bottom-3 repositories, we train the base model with two repo-clusters: a positive cluster (Django plus top-3 Django-like repositories) and a negative cluster (Django plus bottom-3 Django-like repositories), while keeping the training data volume matched at 8k.
\cref{fig:cluster_transfer} shows that the positive cluster indeed outperforms the negative cluster by 8.4\%, confirming the predictive power of the probe for inter-repo learning transferability.
Notably, learning from the positive cluster outperforms the budget-matched Django-only RCL, indicating the potential of cross-repository learning with carefully selected data source.
The probe thus offers a cheap heuristic to build cluster-experts from compatible repositories, e.g., an enterprise's private codebases that share coding practices.

\subsection{Practicality of RCL}





\begin{figure}[t]
    \centering
    \includegraphics[width=0.75\linewidth]{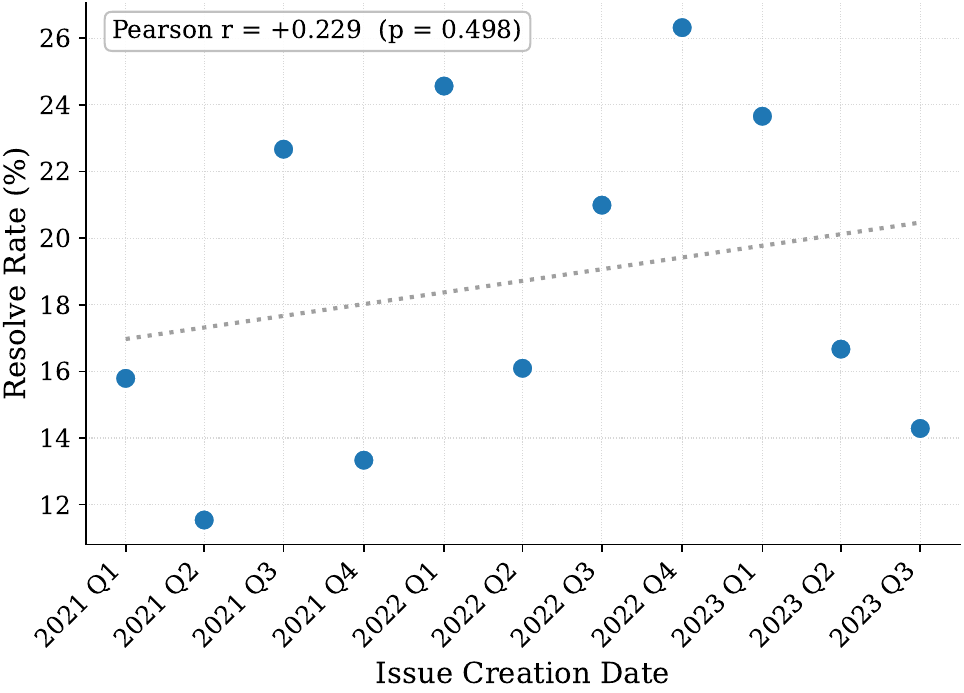}
    \caption{
    \textbf{\model Expert Issue-Resolution Rate vs.\ SWE-bench-Verified Issue Date.} Experts are trained only on repository knowledge before the cutoff date; their issue-resolution rate shows no significant correlation with issue creation date over nearly three years ($p=0.498$).
    }
    \label{fig:perf_vs_time}
\end{figure}

\paragraph{The cost of an RCL expert is quickly amortized}
Training a dedicated expert per repository is a one-time investment, which we estimate is quickly recovered.\footnote{Our estimates assume typical cloud prices at the time of writing: about \$4 per hour for a 141GB-class (H200) GPU and about \$1 per hour for a 48GB GPU.}
The investment is about \$2,500, almost all of which goes to synthesizing training data with the teacher; SFT itself takes about 10 H200 GPU-hours (about \$40, \cref{sec:setup}), and storing an 8GB checkpoint is negligible.
For comparison, TCL relies on the same kind of data synthesis at a much larger scale (e.g., CWM uses 3M trajectories from 3.15K repositories).
In return, every subsequent task becomes cheaper.
At our 48k-token context, a 32B generalist needs a 141GB-class GPU, because its 64GB of weights plus the KV cache exceed an 80GB card, whereas a 4B expert fits on a single 48GB GPU, decodes faster, and serves more concurrent requests.
Together with its shorter trajectories (11.1\% fewer tokens and 23\% fewer turns; \cref{fig:efficiency}), this brings the cost to roughly \$0.1 per task for the expert versus \$1 for the generalist.
The expert thus pays off after about 3,000 tasks and is about $8\times$ cheaper per task afterward; since its accuracy holds for nearly three years (\cref{fig:perf_vs_time}), an actively developed repository easily reaches this volume.

\paragraph{RCL experts are practical to deploy}
A 4B expert also runs on a single 24GB consumer GPU, enabling fully local use in the privacy-sensitive settings that motivate this work (\cref{sec:intro}), where a 32B generalist cannot run.
Serving many experts is practical as well: industrial multi-model serving systems such as NVIDIA Triton and AWS SageMaker multi-model endpoints host thousands of models behind a single endpoint, and 8GB experts are cheap to load and unload.
Moreover, the number of experts need not grow with the repository count: probe-guided cluster experts (\cref{sec:transfer}) let an organization consolidate $N$ repositories into $M \ll N$ experts, reducing both training and serving cost.

\paragraph{RCL experts remain useful over multi-year repository evolution}
\cref{fig:perf_vs_time} studies how a repository expert's performance changes as evaluation instances move away from the training cutoff.
For each target repository, the RCL model is trained only on knowledge available before the cutoff date and evaluated on future SWE-bench Verified issues.
Across nearly three years of post-cutoff issues, issue-resolution performance shows no significant correlation with issue creation time ($p = 0.498$).
This suggests that, for mature repositories and SWE-bench-style tasks, the repository prior learned by RCL does not require frequent retraining to remain useful.



\paragraph{RCL does not necessarily rely on data from realistic PR history}
In our four-unit RCX design, training data of issue resolution and test generation is created based on realistic PR history of the target codebase, while data of the other two units leverage a codebase snapshot without using its PR history.
To test whether RCL needs realistic PR history, we let the teacher model design purely-synthetic problem instances for Django issue resolution on its own, given only a codebase snapshot before the knowledge cutoff date and no historical PR.
Comparing RCL on purely synthetic data and realistic PR data with a matched training data volume at 2k, we find the former does not yield worse performance (pass@1 20.3\%) than the latter (pass@1 18.3\%), indicating that RCL does not necessarily require a realistic PR history and thus is applicable to repositories with sparse PR history. Additionally, we compare issue-resolution performance versus usable PR count for our seven repository experts and found only a weak, statistically-insignificant linear correlation of $r = 0.199$, $p = 0.669$ between the two. This is further evidence that our methods do not depend on strong commit histories.



\section{Related Work}
\label{sec:related}


\paragraph{Software Engineering Tasks and Benchmarks.} Large language model (LLM) agents are increasingly integrated into a wide range of software engineering (SE) tasks. In this work, we focus on three areas most relevant to our study and introduce representative benchmarks: \textit{program repair}~\cite{yang2024swe, gao2022programrepair, li2024hybrid, zhang2025swe, zan2025multi, deng2025swebenchproaiagents, aleithan2024swe, wang2025swe}, \textit{test generation}~\cite{mundler2024swt, huang2025benchmarking, nashid2025issue2test, huang2025benchmarking, jain2025testforge}, and \textit{code question answering}~\cite{liu2021codeqa, hu2024coderepoqa, li2024procqa, chen2025coreqa, hu2025understanding}.
\textit{Program repair} has emerged as a prominent application area, in which an LLM agent analyzes a code repository together with a natural-language issue report, identifies the underlying defects, and automatically proposes candidate patches. These fixes are expected to preserve intended program behavior and pass hidden test suites. A notable benchmark in this space is \textsc{SWE-Bench}~\cite{yang2024swe}, which formalizes real-world, repository-level repair tasks, later \textsc{Mult-SWE-Bench}~\cite{zan2025multi} extends \textsc{SWE-Bench} to multilingual settings.
\textit{Automated test generation} is another key application, where LLM agents synthesize context-aware and comprehensive test cases to improve coverage and enhance validation efficiency~\cite{zhang2024llm, nan2025test, wang2024hits, xue2024llm4fin, yang2024evaluation}. A representative benchmark is \textsc{SWT-Bench}~\cite{mundler2024swt}, a test-generation counterpart derived from \textsc{SWE-Bench}. Finally, \textit{code question answering} supports interactive development workflows by explaining code functionality, identifying potential problems, recommending improvements, and enforcing project-specific conventions. For example, \textsc{CodeQA}~\cite{liu2021codeqa} provides free-form code comprehension questions, while \textsc{SWE-QA}~\cite{peng2025swe} extends beyond \textsc{CodeQA} to operate in realistic, repository-scale environments.

\paragraph{Software Engineering Agents}  To address these real-world SE tasks, a growing number of LLM-based software engineering agents have been proposed. These agents typically rely on general-purpose LLMs as backend models and build specialized workflows or tool-use strategies to tackle specific SE problems. For \textit{program repair agents}~\cite{zhang2024autocoderoverautonomousprogramimprovement, lv2024codeact, wang2024openhands, li2025swe}, representative examples include \textsc{SWE-Agent}~\cite{yang2024swe} and \textsc{Mini-SWE-Agent}~\cite{mini-swe-agent2026}, which adopt a ``reason-then-act'' paradigm with tool invocation, and \textsc{Agentless}~\cite{xia2024agentless}, which follows a static, predefined workflow rather than dynamic reasoning. However, these agents may produce overfitted patches during repair. Building on them, \textsc{Refine}~\cite{pabba2025refine} introduces a general enhancement agent designed to mitigate overfitting and strengthen existing program repair pipelines.
For \textit{test generation agents}~\cite{khatib2025assertflip, kang2023large, ahmed2025heterogeneous}, some approaches adapt program-repair workflows by modifying prompts and reusing their task structure. Beyond these, \textsc{e-Otter++} and \textsc{Issue2Test}~\cite{nashid2025issue2test} leverage execution feedback to iteratively produce higher-quality test cases.
For \textit{code question answering agents}, \texttt{SWE-QA-Agent} performs reasoning and tool-based actions to automatically locate answers, \textsc{CodeAgent}~\cite{tang2024codeagent} introduces autonomous communicative agents for code review, and \textsc{CodeQA-Agent}~\cite{ahmed2024codeqa} combines retrieval-augmented generation (RAG) with LLMs to further improve response accuracy.

\paragraph{Training Specified Models for SE Agents} Although existing agentic workflows demonstrate strong effectiveness in solving SE tasks, most rely on proprietary, commercial LLMs, such as Claude-Sonnet or GPT-4.1, as their backbone. This dependence introduces high computational cost and raises concerns around scalability, accessibility, and on-device deployment. To alleviate these limitations, recent work has begun developing smaller, specialized models to replace heavyweight commercial LLMs~\cite{faircodegenteam2025cwmopenweightsllmresearch, yang2024swe, ma2024lingma,  pan2024training, xie-2025-swefixer, ma2025sorft}. For example, \textsc{Co-PatcheR}~\cite{tang2025copatcher} trains three lightweight models dedicated to bug localization, patch generation, and patch validation for program repair. \textsc{SWE-Exp}~\cite{chen2025swe} distills concise, transferable experience from prior agent trajectories, enabling continuous learning across issue-solving episodes. Similarly, \textsc{SWE-Mirror}~\cite{wang2025swe} extracts the semantic core of real-world issues, mirrors them into a different repository configured with a Gym-style environment, and replays them as verifiable issue-resolution tasks.

\section{Conclusion}

In this work, we introduce Repository-Centric Learning to advance training efficient coding agents,
an approach designed to bridge the inference-time capability gap in Small Language Models by parametrically internalizing the dynamics of target codebases.
Using this approach, our \model-4B model breaks the established scaling barriers,
demonstrating that repository mastery is a distinct, necessary, and practical new dimension that complements general coding capability.

\section*{Limitations}
\label{sec:discuss}

\paragraph{Extensions to the RCL Instantiation}
The RCL implementation in this paper is a SLM student learning from teacher-produced RCX through SFT, due to our limited compute budget.
In the future, RCL can be implemented with other training algorithms, such as on-policy distillation \cite{lu2025onpolicydistillation}, or RLVR which allows agents to learn from self-produced RCX but much more training-compute expensive.
While the RCL-over-TCL advantage persists with a weaker teacher, the absolute gains over the base model also depend on teacher quality, and our scaling evidence beyond 4B is limited to a single 30B student on Django.

\paragraph{Covering More Repository Types}
Due to the limitation of existing accessible and recognized benchmark suite for coding agents, we evaluate RCL on seven mature open-source Python repositories that have the most evaluation instances across our selected benchmarks.
Our JavaScript pilot (\cref{sec:rq1}) covers only one repository with 17 instances, as JavaScript is the only other language for which TCL training data (SWE-Smith) is available.
Future work is needed to evaluate on other types of repositories, such as private/non-open-source ones, less-mature ones, or low-resource-language ones, to confirm a broader generalizability.

\bibliography{main}

\appendix


\section{Details on Repository-Centric Experience Collection}
\label{app:RCX}

Here, we provide more implementation details of our collection of Repository-Centric Experience (RCX).

\subsection{Active Code Analysis for Software Design}
\label{app:active-code-review}

We sample modules, files, and code chunks in a target repository as objects at varying granularity for active code analysis, sharing the underlying idea with seed-conditioned data synthesis \cite{nagarajan2025roll}.
We prioritize objects with richer commit history, as these hotter regions are more likely to be central to the repository's evolution and relevant to future tasks.
Given a target object, the teacher agent is allowed to explore the repository through normal agent actions such as searching for symbols, reading related files, and inspecting tests or usages.
It then produces a structured report describing the object's functionality, design rationale, dependencies, usage patterns, and interactions with the rest of the codebase.
This design is motivated by the limitation of purely syntactic static analysis: many repository-level design signals are encoded in names, comments, conventions, and repeated usage patterns rather than in explicit dependency edges alone.

\subsection{Agentic FIM for Contextual Implementation}

To operationalize the Contextual Implementation unit of RCX, we transform the traditional Fill-in-the-Middle (FIM) objective into an agentic task. Unlike standard FIM, which relies on passive context window stuffing, our Agentic FIM requires the model to actively explore the repository to discover cross-file dependencies.

We utilize static analysis to assist in constructing instances that require repository-level reasoning:
\begin{itemize}[nosep, leftmargin=*]
    \item AST-Aligned Hole Sampling: We parse source files into Abstract Syntax Trees (AST) to select candidate holes aligned with syntactic structures (complete functions or classes) rather than arbitrary text spans.
    \item Dependency Verification: To ensure the task requires cross-file reasoning, we employ a Language Server Protocol (LSP) backend (e.g., Pyright) to resolve symbols within the candidate hole to their definitions in other files.
    \item Hole Classification \& Weighted Sampling: Candidates are classified as Positive (containing resolved cross-file calls) or Negative (local logic only). To maximize learning efficiency, we heavily downsample negative holes and prioritize positive holes that cover a diverse set of repository dependencies.
\end{itemize}
For each selected instance, the target span is removed from a clean repository snapshot and the agent is asked to restore the missing implementation.
The static-analysis results used during construction are not given to the agent; they only ensure that the sampled tasks contain nontrivial repository-level dependencies.
In some instances, a small number of related entities from other files are also removed, forcing the agent to recover a coherent implementation across files rather than completing a purely local span.

\subsection{PR Mirror for Evolutionary Replay}
\label{app:pr-mirror}
To learn from the software evolution of a target repository, we rely on its history of pull requests (PRs) to create issue-resolving tasks which encourage the agent to learn from previous pitfalls in the repository. We adapt the PR Mirror bug generation method from \citet{yang2025swesmith},
whereby changes from previous PRs are undone by an agent in the current (or target) commit of the repository. These bugs introduced by the agent are then validated against a set of tests from the repository's test suite. While \citet{yang2025swesmith} are cautious in their PR selection,
our method strays with the goal of maximizing number of PRs used for bug creation, optimizing gain from this limited but vital resource.

Specifically, \citet{yang2025swesmith} require PRs used for bug generation to have a linked issue with a problem statement, a test patch, and strict limits on number of lines changed. On the other hand, our method utilizes LLM agents for problem statement generation for PRs with no linked issue text (which is the majority); a deterministic test finding algorithm to find the relevant subset of the test suite to validate the bug against instead of relying on a test patch; and have loose restrictions on number of lines changed, leveraging the larger context windows of modern models. As a result, our method results in a roughly \textit{18-fold} increase in task instances per repository as compared with the original PR Mirror method with little to no degradation in task quality.

For repositories with limited usable PR history, we further augment the reality-grounded tasks with synthetic issue-resolution instances generated inside the target repository snapshot.
These synthetic tasks follow the same agentic workflow as PR-history-derived tasks and are used to improve coverage when the historical record alone is too sparse.

\subsection{Test Reproduction for Semantic-Runtime Alignment}
\label{app:sr-align}
As mentioned, we use the same bugs generated by our adapted PR Mirror method for test generation trajectories. The difference being that the prompt used to feed in the task instance to the agent directly explains the workflow for writing reproduction tests. The agent is instructed to either edit existing testing files in the repository or place new ones within the existing testing framework. This way the agent must explore and understand the testing framework for the repository in order to succeed in its task.
Generated tests are retained when they specify the intended semantic correction by failing on the buggy code and passing after the fixed code is restored.
Thus, the same underlying historical fault yields two complementary RCX signals: one trajectory for repairing the behavior and another for formalizing the expected behavior as executable tests.

\end{document}